\documentclass[sigconf]{acmart}

\AtBeginDocument{%
  \providecommand\BibTeX{{%
    \normalfont B\kern-0.5em{\scshape i\kern-0.25em b}\kern-0.8em\TeX}}}

\setcopyright{acmcopyright}
\copyrightyear{2023}
\acmYear{2023}
\acmDOI{XXXXXXX.XXXXXXX}

\acmConference[CSCW '23]{The 26th ACM Conference On Computer-Supported Cooperative Work And Social Computing}{Oct 14--18,
   2023}{Minneapolis, MN}

\usepackage{graphicx}
\usepackage{subcaption}

\begin{document}

%%
%% The "title" command has an optional parameter,
%% allowing the author to define a "short title" to be used in page headers.
\title{FeedbackMap:  a tool for making sense of open-ended survey responses}

%%
%% The "author" command and its associated commands are used to define
%% the authors and their affiliations.
%% Of note is the shared affiliation of the first two authors, and the
%% "authornote" and "authornotemark" commands
%% used to denote shared contribution to the research.

\author{Doug Beeferman}
\affiliation{%
  \institution{Massachusetts Institute of Technology}
  \city{Cambridge, MA}
  \country{USA}}
\email{dougb5@mit.edu}

\author{Nabeel Gillani}
\affiliation{%
  \institution{Northeastern University}
  \city{Boston, MA}
  \country{USA}}
    \email{n.gillani@northeastern.edu}

%%
%% By default, the full list of authors will be used in the page
%% headers. Often, this list is too long, and will overlap
%% other information printed in the page headers. This command allows
%% the author to define a more concise list
%% of authors' names for this purpose.
\renewcommand{\shortauthors}{Beeferman and Gillani}

%%
%% The abstract is a short summary of the work to be presented in the
%% article.
\begin{abstract}
Analyzing open-ended survey responses is a crucial yet challenging task for social scientists, non-profit organizations, and educational institutions, as they often face the trade-off between obtaining rich data and the burden of reading and coding textual responses. This demo introduces FeedbackMap, a web-based tool that uses natural language processing techniques to facilitate the analysis of open-ended survey responses. FeedbackMap lets researchers generate summaries at multiple levels, identify interesting response examples, and visualize the response space through embeddings.   We discuss the importance of examining survey results from multiple perspectives and the potential biases introduced by summarization methods, emphasizing the need for critical evaluation of the representation and omission of respondent voices.
\end{abstract}

%%
%% The code below is generated by the tool at http://dl.acm.org/ccs.cfm.
%% Please copy and paste the code instead of the example below.
%%
% \begin{CCSXML}
% <ccs2012>
%  <concept>
%   <concept_id>10010520.10010553.10010562</concept_id>
%   <concept_desc>Computer systems organization~Embedded systems</concept_desc>
%   <concept_significance>500</concept_significance>
%  </concept>
%  <concept>
%   <concept_id>10010520.10010575.10010755</concept_id>
%   <concept_desc>Computer systems organization~Redundancy</concept_desc>
%   <concept_significance>300</concept_significance>
%  </concept>
%  <concept>
%   <concept_id>10010520.10010553.10010554</concept_id>
%   <concept_desc>Computer systems organization~Robotics</concept_desc>
%   <concept_significance>100</concept_significance>
%  </concept>
%  <concept>
%   <concept_id>10003033.10003083.10003095</concept_id>
%   <concept_desc>Networks~Network reliability</concept_desc>
%   <concept_significance>100</concept_significance>
%  </concept>
% </ccs2012>
% \end{CCSXML}

% \ccsdesc[500]{Computer systems organization~Embedded systems}
% \ccsdesc[300]{Computer systems organization~Redundancy}
% \ccsdesc{Computer systems organization~Robotics}
% \ccsdesc[100]{Networks~Network reliability}

%%
%% Keywords. The author(s) should pick words that accurately describe
%% the work being presented. Separate the keywords with commas.

% \received{20 February 2007}
% \received[revised]{12 March 2009}
% \received[accepted]{5 June 2009}

%%
%% This command processes the author and affiliation and title
%% information and builds the first part of the formatted document.
\maketitle

\section{Introduction}

Open-ended survey questions can give richer information to researchers than closed-ended questions, with lower risks of certain kinds of bias \cite{reja2003open}.  But in deciding whether to add such a question, survey makers face a trade-off between obtaining rich data and the burden of reading and coding textual responses.   We seek to reduce that burden and thereby make it more compelling for surveyors to add questions with free-form textual answers.  We introduce FeedbackMap \footnote{ A public version is available at \href{https://feedbackmap.org}{https://feedbackmap.org} and its source code is available at \href{https://github.com/Plural-Connections/feedbackmap}{https://github.com/Plural-Connections/feedbackmap} for organizations wishing to host it themselves.}, a tool that summarizes a collection of open-ended responses in multiple ways.

Our goal is a kind of multi-document summarization, which is well-studied in the context
of news stories and business communication. 
Survey responses differ from these kinds of document collections in that they are from individuals expressing different perspectives in response to a common question.  A summary of any text loses some of the nuance of the input, but with survey responses this means that individual perspectives may be erased.  A survey summary may systematically prefer the majority response, or exclude certain types of responses due to hidden biases in the summarization model.  Without careful evaluation, the analyst may jump to conclusions that influence actions taken on behalf of the communities they serve.    As such, there are three priorities that guide the development of FeedbackMap:  (1) Give the researcher multiple perspectives on the data; (2) show interactions between open-ended responses and known categorical variables; and (3) connect findings back to individual responses to the survey.  

\section{Related Work}

\subsection{Natural language processing for qualitative analysis}  Qualitative researchers use tools like NVIVO \cite{NVivo} to support manual open-ended analyses, but they are often expensive to access and do little to reduce the time burden of qualitative analyses.  In recent years, researchers have turned to automated methods to gather new insights from open-ended corpora in time-efficient ways~\cite{nelson2020comp,roberts2014stm}.  However, many of the most popular and frequently used methods are also dated in how they operate in settings of data sparsity (like Latent Dirichlet Allocation~\cite{blei2003lda}), rendering them less effective in surfacing nuanced and actionable findings from text corpora.  Even when such tools make use of recent and more powerful methods (like BERTopic~\cite{grootendorst2022bertopic}), they often require writing code to model and analyze bespoke datasets, creating barriers for those who are not trained as engineers and analysts.  Startups are jumping in to fill this void, but may charge fees that make them inaccessible to community organizations with limited budgets.  There appears, then, to be a gap between effectiveness, user-friendliness, and cost in the domain of open-ended survey response analysis. Researchers are beginning to design and prototype platforms to bridge this gap~\cite{goldman2022quad, mellon2022does}, yet more work is needed to make these tools more general and accessible to wider audiences while simultaneously remaining flexible enough to incorporate emerging capabilities from the NLP community. 

\subsection{Biases in open-ended survey response analysis}  
Bias is inherent in any type of voluntary survey, and in particular, open-ended survey questions.  A recent study by Pew Research found that women, younger adults, Hispanic and Black adults, and individuals with less formal education were less likely to answer open-ended survey questions~\cite{pew2023survey}, reflecting findings from prior work~\cite{andrews2005survey}.  Of course, even when people do respond, feelings of ``question threat'' may contribute to biased or misleading responses~\cite{bradburn1978threat}.

Bias stems both from \textit{who responds} as well as \textit{how responses are analyzed}.  Practices like concept mapping~\cite{jackson2002concept} and computing inter-coder reliability across multiple coders for the same content~\cite{oconnor2020icr} seek to account for researcher biases in the analysis process.  FeedbackMap's multiple data presentations seek to mitigate biases that may be reinforced through a single analysis frame, yet the extent to which this succeeds, and how it may help address issues of response biases further upstream in the feedback-giving process---remain important open questions for our work.

\section{System Overview}

FeedbackMap is a Web application implemented using the Streamlit framework \cite{Streamlit}.  Organizations that host the tool can customize the language models (LMs) to use for writing summaries and computing embeddings. LMs may be API-based, such as the GPT3 model from the OpenAI API \cite{OpenAI_API}--the choice for summaries in our publicly deployed instance of the tool--or a local transformer-based model from PyTorch-transformers \cite{wolf2020transformers}.   Other customizations include the types of summarization available to the end user and their corresponding LM prompts.  The tabs shown to the end user as they use the Web application are described below.

\subsection{Welcome tab}
The Welcome tab invites the researcher to upload a comma-separated value (CSV) file containing the results of their survey.   The tool is designed to work best with the format that is produced by Google Forms, but it will try to accommodate any CSV or JSONL-formatted file.  While the input file may be arbitrarily large, the records will be randomly sampled for analysis if they exceed a threshold (5000 rows, in the case of our publicly deployed instance.)

\subsection{Summary tab}
The Summary tab gives an overview of the selected data file and asks the researcher to pick one of the open-ended questions in the data to analyze.   Here, by “open-ended” we mean that the survey asked for a free-form textual response.  While we can’t be certain of the question type from a schema-less CSV file, in practice it’s usually easy to tell:  FeedbackMap infers which columns of data are open-ended and which are categorical by analyzing the distribution of answer values for each question.  The open-ended questions are shown alongside the rate of nonempty responses for the question, while the categorical questions are shown alongside information about the observed value distribution.  For categorical questions identified as multi-select, each value is counted separately.   Menus next to the categorical questions let the user constrain the analysis (for example, to consider only the responses from one U.S. state) prior to clicking on an open-ended question.

\begin{figure}[h]
    \centering
    \begin{subfigure}{0.4\textwidth}
        \centering
        \frame{\includegraphics[width=4cm]{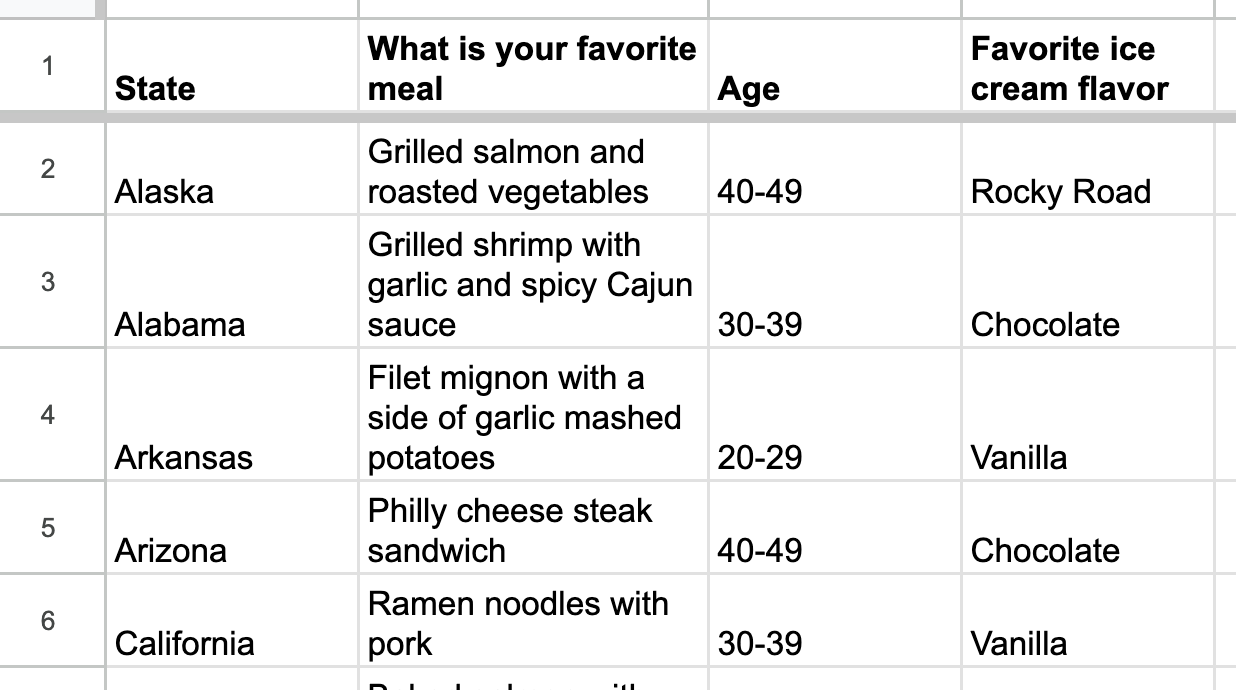}}
        \label{fig:inputdata}
    \end{subfigure}
    \hspace{1cm}
    \begin{subfigure}{0.4\textwidth}
        \centering
        \frame{\includegraphics[width=4cm]{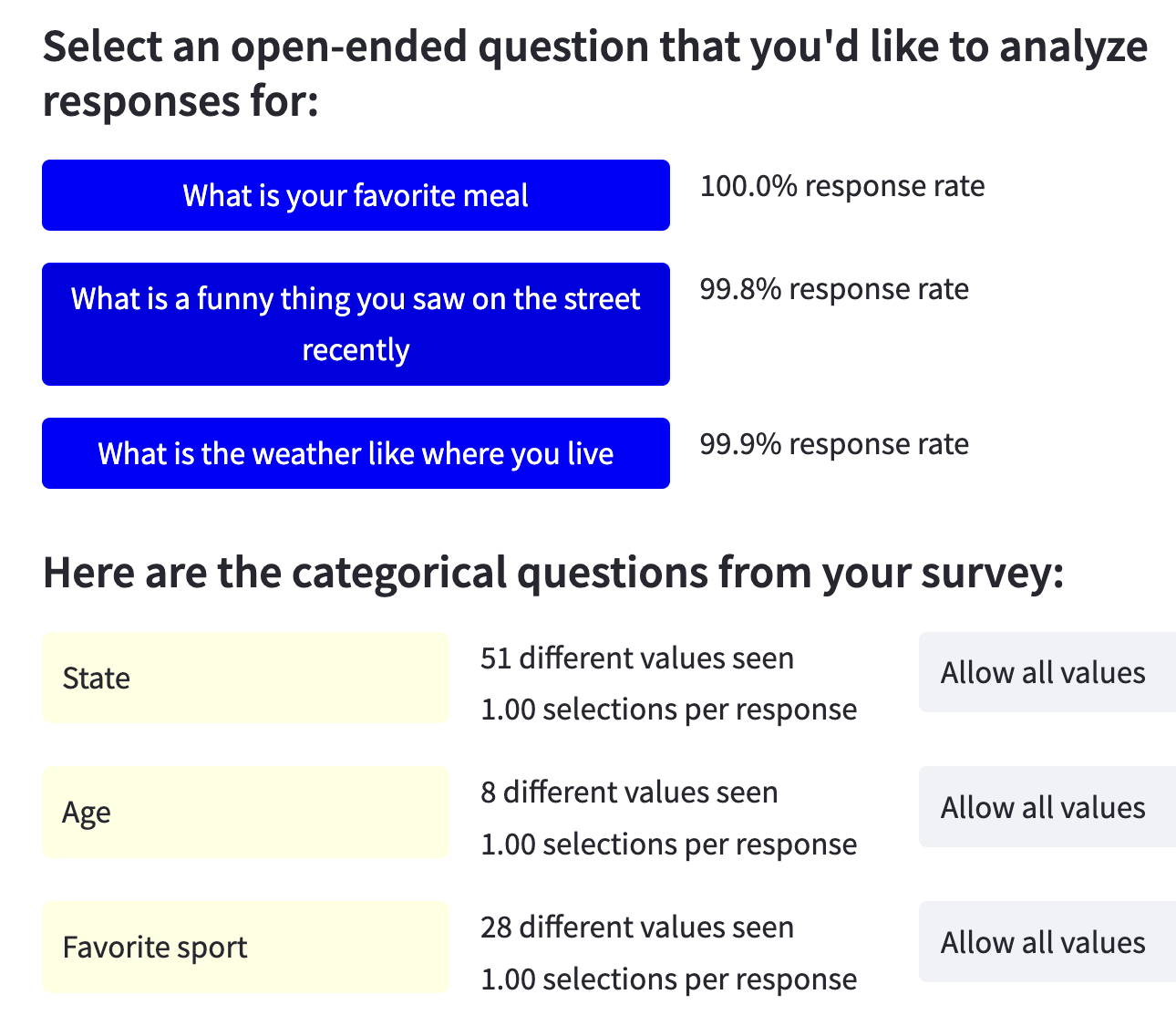}}
        \label{fig:summarytab}
    \end{subfigure}
    \caption{An excerpt of the input file (left) and summary tab (right) for our synthetic survey response data.  The dataset consists of 1,020 GPT-4-generated responses to questions about age, state of residence, favorite meal, and the current weather, among others.  The script to generate this data and its output are available in the demo repository.}
    \label{fig:input-and-summary}
\end{figure}

Figure \ref{fig:input-and-summary} shows an excerpt of a synthetic survey data file, and the corresponding Summary tab that FeedbackMap displays for the file.

\subsection{Analysis tab}
The analysis tab contains the key offerings of the tool, letting the user explore the open-ended responses to a question of interest in various ways, each way contained within a collapsible subsection.   Figure \ref{fig:analysistab} shows the subsections of the analysis tab for our hypothetical input file.   We discuss these in more detail below:

\begin{figure*}

    \begin{subfigure}[b]{0.45\textwidth}
        \centering
        \frame{\includegraphics[width=\textwidth,height=2.6cm]{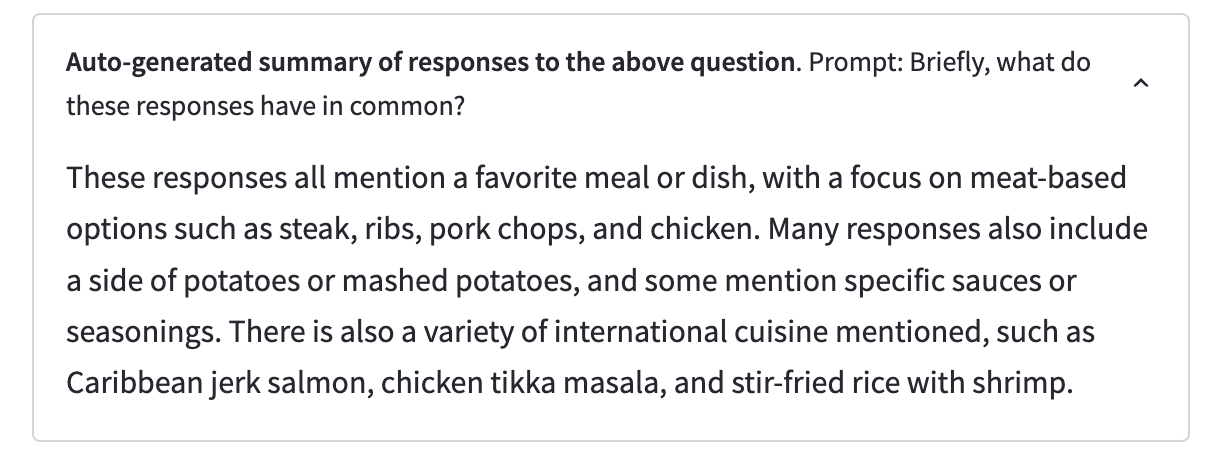}}
        \caption{Top-level abstractive summmary}
        \label{fig:autosummary}
    \end{subfigure}
    \hspace{0.3cm}
    \begin{subfigure}[b]{0.45\textwidth}
        \centering
        \frame{\includegraphics[width=\textwidth,height=2.6cm]{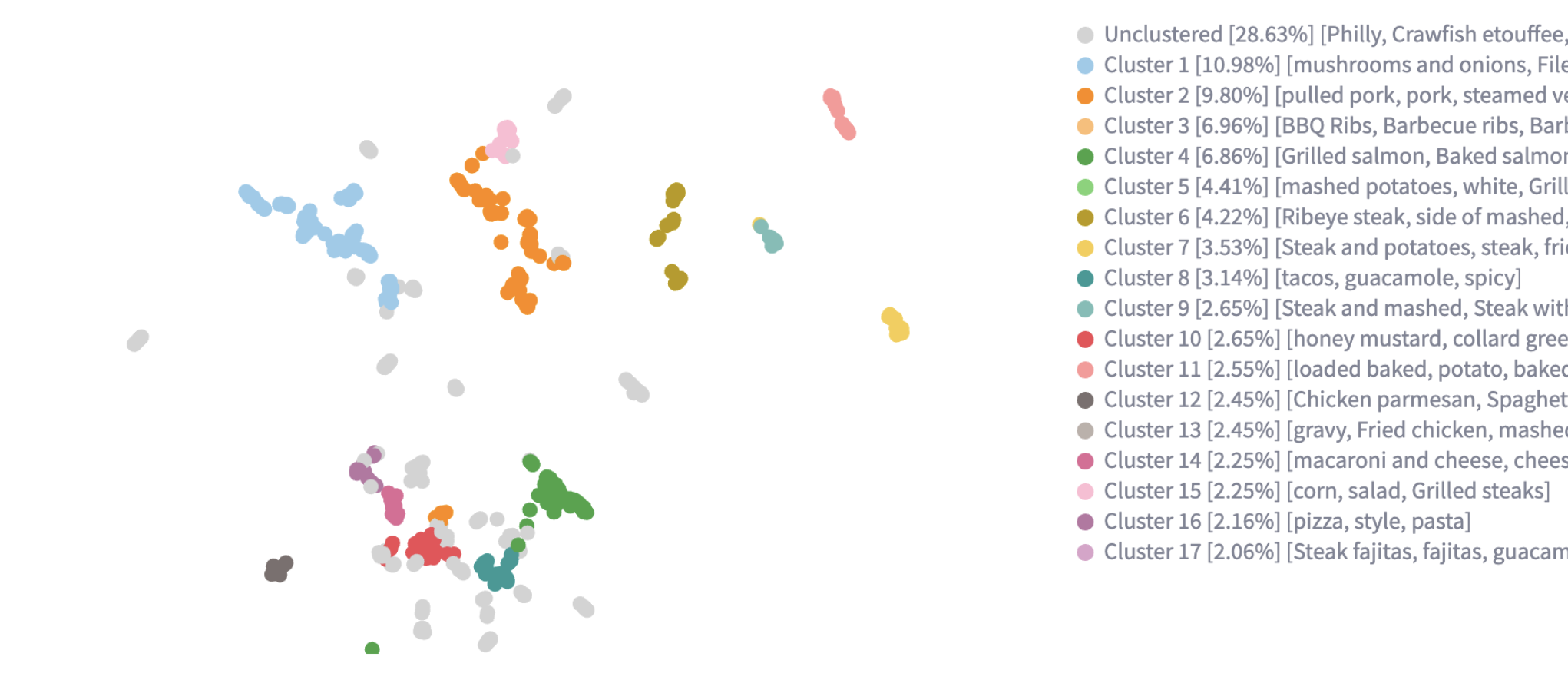}}
        \caption{Topic scatterplot}
        \label{fig:scatterplot}
    \end{subfigure}

    \begin{subfigure}[b]{0.45\textwidth}
        \centering
        \frame{\includegraphics[width=\textwidth,height=2.6cm]{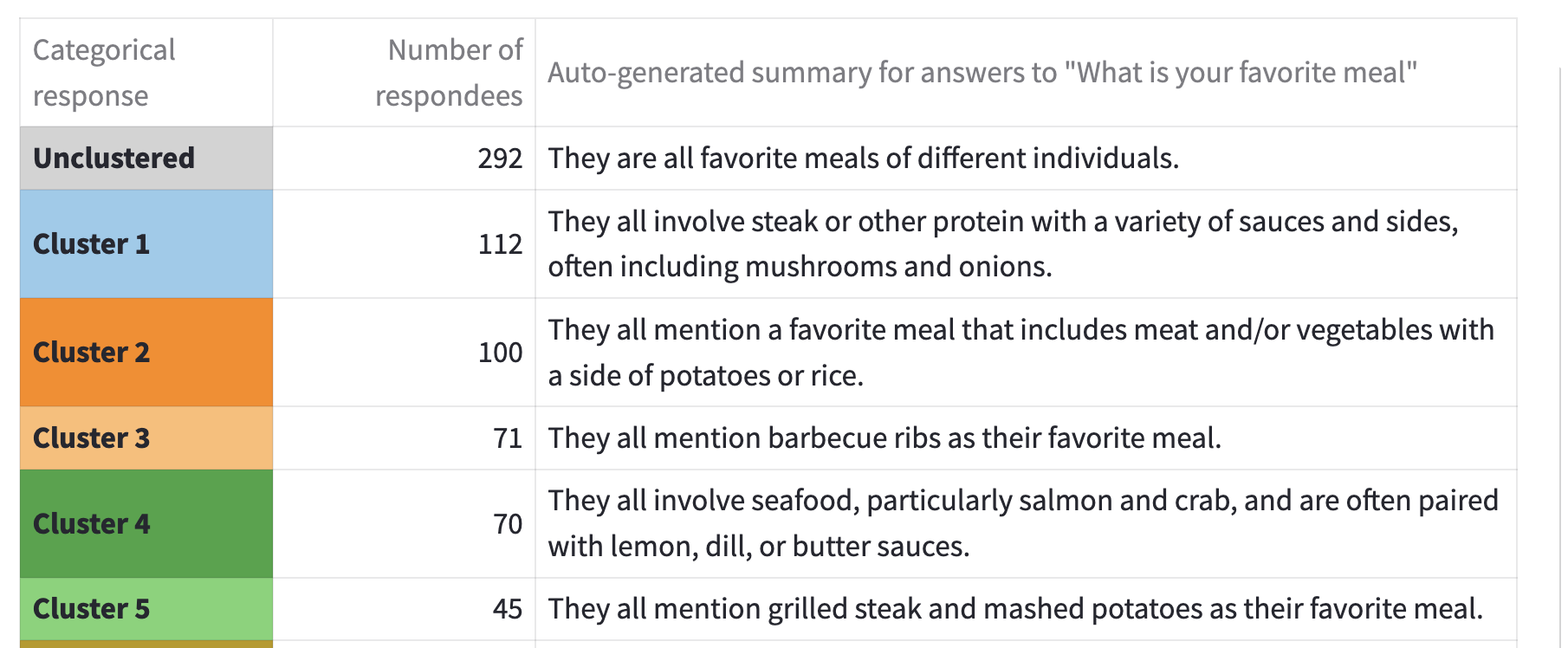}}
        \caption{Cluster summaries (clustering by response text)}
        \label{fig:categorybreakdown}
    \end{subfigure}
    \hspace{0.2cm}
    \begin{subfigure}[b]{0.45\textwidth}
        \centering
        \frame{\includegraphics[width=\textwidth,height=2.6cm]{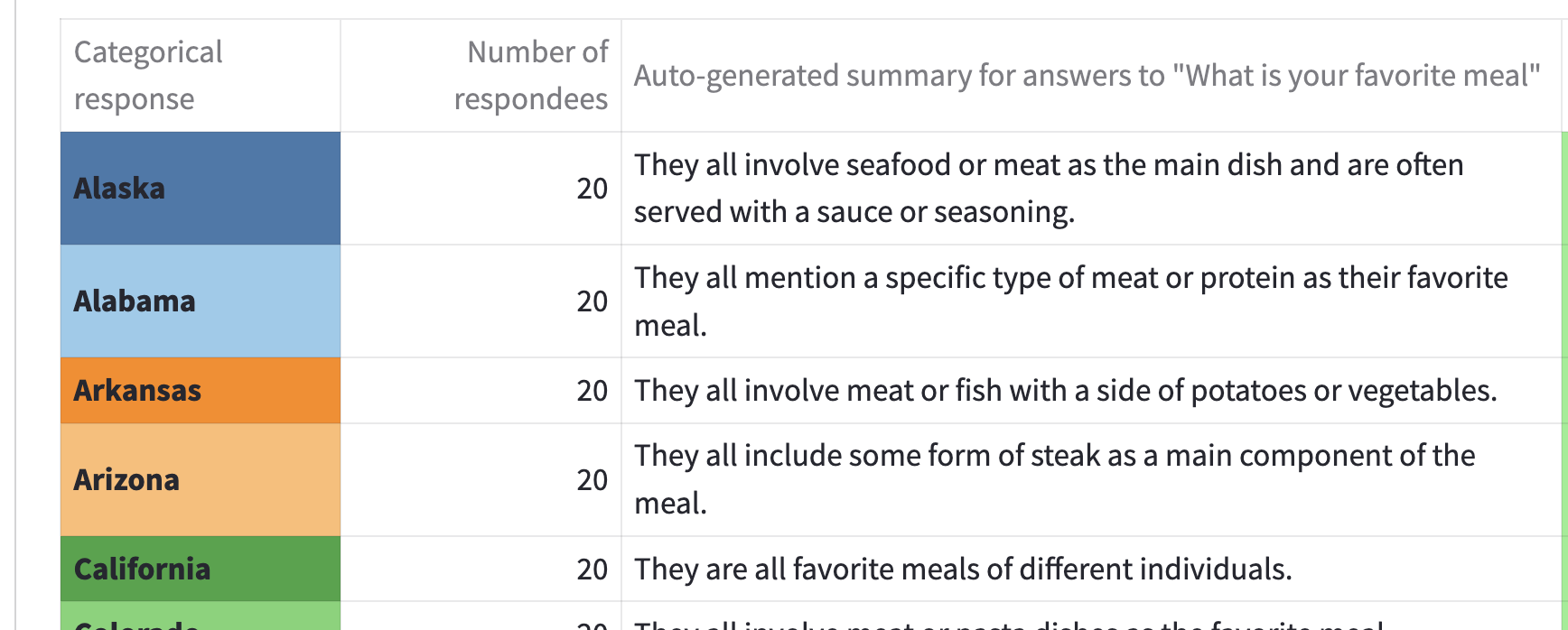}}
        \caption{Cluster summaries (clustering by U.S. state)}
        \label{fig:categorybreakdown}
    \end{subfigure}
    \caption{Some elements of the Analysis tab when the user selects the question "What is your favorite meal?"}
    \label{fig:analysistab}
\end{figure*}

\subsubsection{Top-level abstractive summary} 
This section shows a top-level summary of the responses to the selected open-ended question.  It is generated using an LM and prompt that may be controlled in the code configuration.  The prompt to the LM is a random sample of responses (sample size chosen to maximize the use of the LM's input context window), followed by an instructional statement such as the default, "Briefly, what do these responses have in common?"

\subsubsection{Topic scatterplot}
This section shows a two-dimensional interactive scatterplot that organizes the individual responses by topic, such that responses about similar topics are close to each other.  Users may read the individual responses by hovering over one of the points.   This section uses a familiar pipeline of techniques (used by, e.g., BERTopic~\cite{grootendorst2022bertopic}) to go from the raw text to the scatterplot:  namely, it computes the embedding for each response according to a sentence embedding model, and then projects the set to 2 dimensions.   By default, the points are clustered and color-coded according to the result of a clustering algorithm applied to these same embeddings.   We use UMAP \cite{mcinnes2018umap} for projection and HDBSCAN \cite{mcinnes2017hdbscan} for clustering.   The user may choose to override this "auto-clustering" and instead group by one of the categorical variables in the survey data.   Labels for the clusters are determined by terms in the responses that have high pointwise mutual information with respect to the other clusters.

\subsubsection{Interesting examples}
Here FeedbackMap uses the LM to ask for noteworthy responses.   As in the top-level summary, this happens by prompting the LM with a random selection of responses, followed by a fixed instructional prompt.  The default instructional portion is “What are 3 interesting responses and why?”, which elicits rationales for the LM’s choices.  A “Pick again” prompt allows the user to rerun the generation, based on a new random sample.\footnote{There is no guarantee that the LM generates examples that are actually in the sample we include in the prompts;  so far we have not seen a case of a hallucinated example, but this is certainly possible and should be part of any evaluation of the tool.}

\subsubsection{Cluster summaries}
This section shows an LM-generated summary for each cluster, color-coded to match the scatterplot.

\subsubsection{Top words and phrases}
We identify words and collocations (multi-word terms) that are frequent in the data and show how they interact with the selected category in tabular form.  Columns of the table correspond to categorical values and can be sorted to surface terms that are highly associated with the category.

\section{Discussion and evaluation plan}

Since deploying FeedbackMap we have shared it with three users, including a school district administrator, a non-profit founder, and a political scientist, all of whom had suitable data sets for it. Reactions were positive, and varied with respect to which features were seen as most useful, with the top-level and category-specific summaries being highlights for all three users.   
We hope to run a more rigorous field study of the tool with these users and others, pending IRB approval.  We also plan to measure the impact of survey summarization on users with respect to the kinds of knowledge they impart and conceal about the underlying data set.

\bibliographystyle{ACM-Reference-Format}
\bibliography{sample-base}

\end{document}